\documentclass[11pt,letterpaper]{article}
\usepackage{ijcnlp2017}
\usepackage{times}
\usepackage{latexsym}
\usepackage{url}
\usepackage{bm}
\usepackage{latexsym}
\usepackage{microtype}
\usepackage{graphicx}
\usepackage{wasysym}
\usepackage{multirow}
\usepackage[hang,flushmargin]{footmisc}

\usepackage{amsfonts}
\ijcnlpfinalcopy

\def\ijcnlppaperid{***}

\title{Emotion Detection on TV Show Transcripts with\\Sequence-based Convolutional Neural Networks}

\author{Sayyed M. Zahiri \\ Mathematics and Computer Science \\  Emory University \\ Atlanta, GA 30322, USA \\ \tt{sayyed.zahiri@emory.edu}
         \And
         Jinho D. Choi \\ Mathematics and Computer Science \\  Emory University \\ Atlanta, GA 30322, USA \\ \tt{jinho.choi@emory.edu}}

\date{}

\begin{document}

\maketitle

\begin{abstract}
While there have been significant advances in detecting emotions from speech and image recognition, emotion detection on text is still under-explored and remained as an active research field.
This paper introduces a corpus for text-based emotion detection on multiparty dialogue as well as deep neural models that outperform the existing approaches for document classification.
We first present a new corpus that provides annotation of seven emotions on consecutive utterances in dialogues extracted from the show, \textit{Friends}.
We then suggest four types of sequence-based convolutional neural network models with attention that leverage the sequence information encapsulated in dialogue.
Our best model shows the accuracies of 37.9\% and 54\% for fine- and coarse-grained emotions, respectively.
Given the difficulty of this task, this is promising.
\end{abstract}

\section{Introduction}

Human emotions have been widely studied in the realm of psychological and behavioral sciences as well as computer science~\cite{strapparava2008learning}. 
A wide variety of researches have been conducted in detecting emotions from facial expressions and audio waves~\cite{yu2001emotion,zeng2006spontaneous,lucey2010extended}.
The recent advent of natural language processing and machine learning has made the task of emotion detection on text possible, yet since emotions are not necessarily conveyed on text, quantifying different types of emotions using only text is generally challenging.
%Given that text-based emotion detection has numerous applications, many computational linguistics  have recently tackled this task with different approaches. 

Another challenging aspect about this task is due to the lack of annotated datasets. 
There are few publicly available datasets~\cite{strapparava2007semeval,alm:08a,MohammadB17wassa,buechel-hahn:2017:EACLshort}.
However, in order to further explore the feasibility of text-based emotion detection on dialogue, a more comprehensive dataset would be desired.
This paper presents a new corpus that comprises transcripts of the TV show, \textit{Friends}, where each utterance is annotated with one of the seven emotions: \textit{sad}, \textit{mad}, \textit{scared}, \textit{powerful}, \textit{peaceful}, \textit{joyful}, and \textit{neutral}.
Several annotation tasks are conducted through crowdsourcing for the maintenance of a high quality dataset.
Dialogues from these transcripts include disfluency, slangs, metaphors, humors, etc., which make this task even more challenging.
To the best of our knowledge, this is the largest text-based corpus providing fine-grained emotions for such long sequences of consecutive utterances in multiparty dialogue. 
%Currently, there are few publicly available copora such as the Affect data~\cite{alm:08a}, the ISEAR databank,\footnote{\url{emotion-research.net/toolbox/toolboxdatabase.2006-10-13.2581092615}}, the EmoBank~\cite{buechel-hahn:2017:EACLshort}, the SemEval'07 Task 14~\cite{strapparava2007semeval}, and the WASSA'17 Task 1~\cite{MohammadB17wassa} datasets.
%Annotating and analyzing texts coming directly from spoken dialogues are challenging in general. Mostly these dialogues contain slangs, word repetition, disfluencies, words specific to special group of people, time or origins and so on ; therefore, it usually requires extensive amount of preprocessing to prepare and clean the text before applying any NLP techniques. The annotated corpus we introduce in this paper is based on multiparty dialogues. 

Convolutional neural networks (CNN) have been popular for several tasks on document classification.
One of the major advantages of CNN is found in its capability of extensive feature extraction through deep-layered convolutions.
Nonetheless, CNN are often not used for sequence modeling~\cite{waibel1989phoneme,lecun1995convolutional,gehring2017convolutional} because their basic architecture does not take previous sequence information into account. 
One common approach to alleviate this issue is using recurrent neural networks~\cite{sutskever2014sequence,liu2016recurrent}.
However, RNNs typically perform slower and require more training data to avoid overfitting. % as they are prone to overfit faster than CNN.
To exploit the sequence information embedded in our corpus yet to employ the advantages of CNN, sequenced-based CNN (SCNN) are proposed along with attention mechanisms, which guide CNN to fuse features form the current state with features from the previous states. 
The contributions of this research are summarized as follows:

\begin{table*}[htbp!]
\centering\resizebox{\textwidth}{!}{
\begin{tabular}{c||l||c|c|c|c}
\bf Speaker & \multicolumn{1}{c||}{\bf Utterance} & \bf A1 & \bf A2 & \bf A3 & \bf A4 \\
\hline\hline
Monica & He is so cute . So , where did you guys grow up ? & Peaceful & Joyful & Joyful & Joyful \\
Angela & Brooklyn Heights . & Neutral & Neutral & Neutral & Neutral \\
Bob    & Cleveland . & Neutral & Neutral & Neutral & Neutral \\
Monica & How , how did that happen ? & Peaceful & Scared & Neutral & Neutral \\
Joey   & Oh my god . & Joyful & Sad & Scared & Scared \\
Monica & What ? & Neutral & Neutral & Neutral & Neutral \\
Joey   & I suddenly had the feeling that I was falling . But I 'm not . & Scared & Scared & Scared & Scared
\end{tabular}}
\caption{An example of the emotion annotation. A\#: annotation from 4 different crowd workers.}
\label{tbl:emotion_annotation}
\end{table*}

\begin{itemize}
\item We create a new corpus providing fine-grained emotion annotation on dialogue and give thorough corpus analytics (Section~\ref{dataset}) .
\item We introduce several sequence-based convolution neural network models with attention to facilitate sequential dependencies among utterances. (Section~\ref{approach}).
\item We give both quantitative and qualitative analysis that show the advantages of SCNN over the basic CNN as well as the advances in the attention mechanisms (Section~\ref{exp}).
\end{itemize}

\section{Related Work}

%In this section we briefly summarize previous researches relevant to text-based emotion detection as well as sequence modeling specifically on dialogue using CNN.

\subsection{Text-based Emotion Detection}
Text-based emotion detection is on its early stage in natural language processing although it has recently drawn lots of attention.
There are three common methods researchers have employed for detecting emotions from text: keyword-based, learning-based and hybrids of those two.
In the first method, classification is done by aid of emotional keywords.
\newcite{strapparava2004wordnet} categorized emotions by mapping keywords in sentences into lexical representation of affective concepts.
\newcite{chaumartin2007upar7} did emotion detection on news headlines.
The performance of these keyword-based approaches had been rather unsatisfactory due to the fact that the semantics of the keywords heavily depend on the contexts, and it is significantly affected by the absence of those keywords~\cite{shaheen2014emotion}

Two types of machine learning approaches have been used for the second method, supervised approaches where training examples are used to classify documents into emotional categories, and unsupervised approaches where statistical measures are used to capture the semantic dependencies between words and infer their relevant emotional categories.
\newcite{chaffar2011using} detected emotions from several corpora collected from blog, news headline, and fairy tale using a supervised approach. 
\newcite{hajar2016using} used YouTube comments and developed an unsupervised learning algorithm to detect emotions from the comments.
Their approach gave comparative performance to supervised approaches such as \newcite{chaffar2011using} that support vector machines were employed for statistical learning.

The hybrid method attempts to take advantages from both the keyword-based and learning methods.
\newcite{seol2008emotion} used an ensemble of a keyword-based approach and knowledge-based artificial neural networks to classify emotions in drama, novel, and public web diary.
Hybrid approaches generally perform well although their architectures tend to be complicated for the replication.

\subsection{Sequence Modeling on Dialogue}

The tasks of state tracking and dialogue act classification are ongoing fields of research similar to our task.
\newcite{shi2016multichannel} proposed multi-channel CNN for cross-language dialogue tracking. 
\newcite{dufour2016tracking} introduced a topic model that considered all information included in sub-dialogues to track the dialogue states introduced by the DSTC5 challenge~\cite{kim2016fifth}.
\newcite{stolcke2000dialogue} proposed a statistical approach to model dialogue act on human-to-human telephone conversations.

\subsection{Attention in Neural Networks}

Attention mechanism has been widely employed in the field of computer vision~\cite{mnih2014recurrent, xu1502show} and recently become popular in natural language processing as well.
In particular, incorporating attention mechanism has achieved the state-of-the-art performance in machine translation and question answering tasks~\cite{bahdanau2014neural, hermann2015teaching,dos2016attentive}.
Our attention mechanism is distinguished from the previous work which perform attention on two statistic embeddings whereas our approach puts attention on static and dynamically generated embeddings.

%\begin{table*}[htbp!]
%\centering\small
%\begin{tabular}{c||l||c|c|c|c}
%\bf Speaker & \multicolumn{1}{c||}{\bf Utterance} & \bf A1 & \bf A2 & \bf A3 & \bf A4 \\
%\hline\hline
%Monica & He is so cute . So , where did you guys grow up ? & Peaceful & Joyful & Joyful & Joyful \\
%Angela & Brooklyn Heights . & Neutral & Neutral & Neutral & Neutral \\
%Bob    & Cleveland . & Neutral & Neutral & Neutral & Neutral \\
%Monica & How , how did that happen ? & Peaceful & Scared & Neutral & Neutral \\
%Joey   & Oh my god . & Joyful & Sad & Scared & Scared \\
%Monica & What ? & Neutral & Neutral & Neutral & Neutral \\
%Joey   & I suddenly had the feeling that I was falling . But I 'm not . & Scared & Scared & Scared & Scared \\
%
%\end{tabular}
%\caption{An example of the emotion annotation. A\#: annotation from 4 different crowd workers.}
%\label{tbl:emotion_annotation}
%\end{table*}

\section{Corpus}
\label{dataset}

The Character Mining project provides transcripts of the TV show, \textit{Friends}; transcripts from all seasons of the show are publicly available in JSON.\footnote{\url{nlp.mathcs.emory.edu/character-mining}} % where the first two seasons are annotated for the task of character identification~\cite{chen:16a}.
Each season consists of episodes, each episode contains scenes, each scene includes utterances, where each utterance gives the speaker information.
For this research, we take transcripts from the first four seasons and create a corpus by adding another layer of annotation with emotions.
As a result, our corpus comprises 97 episodes, 897 scenes, and 12,606 utterances, where each utterance is annotated with one of the seven emotions borrowed from the six primary emotions in the \newcite{willcox1982feeling}'s feeling wheel, \textit{sad}, \textit{mad}, \textit{scared}, \textit{powerful}, \textit{peaceful}, \textit{joyful}, and a default emotion of \textit{neutral}. 
Table~\ref{tbl:emotion_annotation} describes a scene containing seven utterances and their corresponding annotation from crowdsourcing. 
 
\subsection{Crowdsourcing}
\label{crowd}

Pioneered by \newcite{snow2008cheap}, crowdsourcing has been widely used for the creation of many corpora in natural language processing.
Our annotation tasks are conducted on the Amazon Mechanical Turk.
Each MTurk HIT shows a scene, where each utterance in the scene is annotated by four crowd workers who are asked to choose the most relevant emotion associated with that utterance.
To assign a suitable budget for each HIT, the corpus is divided into four batches, where all scenes in each batch are restricted to the [5, 10), [11, 15), [15, 20), [20, 25] number of utterances and are budgeted to 10, 13, 17, 20 cents per HIT, respectively.
%This allowed to disregard scenes that are too short to have enough contextual information or too long scenes that may requires a significant amount of time for annotating.
Each HIT takes about 2.5 minutes on average, and the entire annotation costs about \$680. The annotation quality for 20\% of each HIT is checked manually and those HITs with poor quality are re-annotated.

\begin{table}[htbp!]
\centering\resizebox{\columnwidth}{!}{
\begin{tabular}{l||r|r|r|r||r}
\multicolumn{1}{c||}{\bf Type} & \multicolumn{1}{c|}{\bf B1} & \multicolumn{1}{c|}{\bf B2} & \multicolumn{1}{c|}{\bf B3} & \multicolumn{1}{c||}{\bf B4} & \multicolumn{1}{c}{\bf Total} \\
\hline\hline
Utterances & 1,742 & 2,988 & 3,662 & 4,214 & 12,606 \\
Scenes     &   241 &   249 &   216 &   191 &    897 \\
\end{tabular}}
\caption{The total number of utterances and scenes in each annotation batch. B\#: the batch number.}
\label{tbl:annotation-batch-distributions}
\end{table}

\subsection{Inter-Annotator Agreement}

Two kinds of measurements are used to evaluate the inter-annotator agreement (Table~\ref{tbl:inter-annotator-agreement}).
First, Cohen's kappa is used to measure the agreement between two annotators whereas Fleiss' kappa is used for three and four annotators.
Second, the partial agreement (an agreement between any pair of annotators) is measured to illustrate the improvement from a fewer to a greater number of crowd workers.
Among all kinds of annotator groups, the kappa scores around 14\% are achieved.
Such low scores are rather expected because emotion detection is highly subjective so that annotators often judge different emotions that are all acceptable for the same utterance.
This may also be attributed to the limitation of our dataset; a higher kappa score could be achieved if the annotators were provided with a multimodal dataset (e.g., text, speech, image).
%Furthermore, some utterances may comprise more than one emotion such it is possible to have a mixture of two or even more emotions.

\begin{table}[htbp!]
\centering\resizebox{0.8\columnwidth}{!}{
\begin{tabular}{c||c|c|c}
\bf \# of annotators & \bf 2 & \bf 3 & \bf 4 \\
\hline\hline
Kappa   & 14.18 & 14.21 & 14.34 \\
Partial & 29.40 & 62.20 & 85.09 \\
\end{tabular}}
\caption{Inter-annotator agreement (in \%).}
\label{tbl:inter-annotator-agreement}
\end{table}

\noindent While the kappa scores are not so much compelling, the partial agreement scores show more promising results.
The impact of a greater number of annotators is clearly illustrated in this measurement; over 70\% of the annotation have no agreement with only two annotators, whereas 85\% of the annotation find some agreement with four annotators.
This implies that it is possible to improve the annotation quality by adding more annotators.

It is worth mentioning that crowd workers were asked to choose from 37 emotions for the first 25\% of the annotation, which comprised 36 secondary emotions from \newcite{willcox1982feeling} and the \textit{neutral} emotion.
However, vast disagreements were observed for this annotation, resulting Cohen's kappa score of 0.8\%.
Thus, we proceeded with the seven emotions described above, hoping to go back and complete the annotation for fine-grained emotions later. 

%\begin{table}[htbp!]
%\centering\small
%\begin{tabular}{c||r|r|r|r|r|r|r||r}
% & \multicolumn{1}{c||}{\bf M} & \multicolumn{1}{c|}{\bf S} & \multicolumn{1}{c|}{\bf J} & \multicolumn{1}{c|}{\bf W} & \multicolumn{1}{c|}{\bf P} & \multicolumn{1}{c|}{\bf D} & \multicolumn{1}{c||}{\bf N} & \multicolumn{1}{c}{\bf T} \\
%\hline\hline
%\bf M &     0 &     0 &     9 &     0 &    12 &    20 &     0 &    41 \\\hline
%\bf S &    47 &     0 &    40 &    17 &    19 &    45 &    63 &   231 \\\hline
%\bf J &    15 &     0 &     0 &     0 &    86 &    11 &     0 &   112 \\\hline
%\bf W &    40 &     0 &    73 &     0 &    28 &     8 &     0 &   149 \\\hline
%\bf P &     0 &     0 &     0 &     0 &     0 &     8 &     0 &     8 \\\hline
%\bf D &     0 &     0 &     0 &     0 &     0 &     3 &     0 &     3 \\\hline
%\bf N &    54 &    98 &   261 &    83 &   107 &    37 &     0 &   640 \\
%\hline\hline
%\bf T &   156 &    98 &   383 &   100 &   252 &   132 &    63 & \multicolumn{1}{c}{-} \\
%\end{tabular}
%\caption{Confusion matrix for the primary emotions. M: Mad ,S: Scared, J: Joyful, W: Powerful, P:Peaceful, D: Sad, N: Neutral}
%\label{confusion}
%\end{table}
%
%\noindent   Table \ref{confusion} indicates raw counts of number of times the emotions were confused by the annotators. Table \ref{confusion} suggests that mainly all the emotions got confused by neutral, in particular joyful confused the most with neutral. 

\subsection{Voting and Ranking}
\label{ssec:voting-ranking}

We propose a voting/ranking scheme that allows to assign appropriate labels to the utterances with disagreed annotation.
Given the quadruple annotation, we first divide the dataset into five-folds (Table~\ref{tbl:voting-scheme}):

\begin{table}[htbp!]
\centering\resizebox{\columnwidth}{!}{
\begin{tabular}{l||r|r}
\multicolumn{1}{c||}{\bf Fold} & \multicolumn{1}{c|}{\bf Count} & \multicolumn{1}{c}{\bf Ratio}\\
\hline\hline
$(a_1 = a_2 = a_3 = a_4)$                                      &   778 &  6.17\\
$(a_1 = a_2 = a_3) \wedge (a_1 \neq a_4)$                      & 5,774 & 45.80\\
$(a_1 = a_2) \wedge (a_1 \neq a_3) \wedge (a_1 \neq a_4)$      & 2,991 & 23.73\\
$(a_1 = a_2) \wedge (a_3 = a_4)$                               & 1,879 & 14.91\\
$\forall_{i, j \in [1,4]}.\: (a_i \neq a_j) \wedge (i \neq j)$ & 1,184 &  9.39\\
\end{tabular}}
\caption{Folds with their utterance counts and ratios (in \%) for voting/ranking. $a_{1..4}$: annotation 1..4.}
\label{tbl:voting-scheme}
\end{table}

\noindent For the first three folds, the annotation coming from the majority vote is considered the gold label ($a_1$).
The least absolute error (LAE) is then measured for each annotator by comparing one's annotation to the gold labels.
For the last two folds, annotation generated by the annotator with the minimum LAE is chosen as gold, which is reasonable since those annotators generally produce higher quality annotation.
With this scheme, 75.5\% of the dataset can be deterministically assigned with gold labels from voting and the rest can be assigned by ranking.

\subsection{Analysis}

Table~\ref{tbl:emotion-dist} shows the distribution of all emotions in our corpus.
The two most dominant emotions, \textit{neutral} and \textit{joyful}, together yield over 50\% of the dataset, which does not seem to be balanced although it is understandable given the nature of this show, that is a \textit{comedy}.
However, when the coarse-grained emotions are considered, \textit{positive}, \textit{negative}, and \textit{neutral}, they yield about 40\%, 30\%, and 30\% respectively, which gives a more balanced distribution.
The last column shows the ratio of annotation that all four annotators agree.
Only around 1\% of the annotation shows complete agreement for \textit{peaceful} and \textit{powerful}, which is reasonable since these emotions often get confused with \textit{neutral}. 

\begin{table}[htbp!]
\centering\resizebox{0.9\columnwidth}{!}{
\begin{tabular}{c|c||r|c}
\bf 3 Emotions & \bf 7 Emotions & \bf Ratio & \bf 4-Agree \\
\hline\hline
Neutral & Neutral & 29.95 & 7.8 \\
\hline
\multirow{3}{*}{Positive} & Joyful   & 21.85 & 9.4 \\
                    & Peaceful &  9.44 & 1.0 \\
                    & Powerful &  8.43 & 0.8 \\
\hline
\multirow{3}{*}{Negative} & Scared   & 13.06 & 3.8 \\
                    & Mad      & 10.57 & 7.7 \\ 
                    & Sad      &  6.70 & 4.3 \\
\end{tabular}}
\caption{Emotion distribution ratios and the complete agreement ratios (in \%).}
\label{tbl:emotion-dist}
\end{table}

\begin{figure}[htbp!]
\includegraphics[width=\columnwidth]{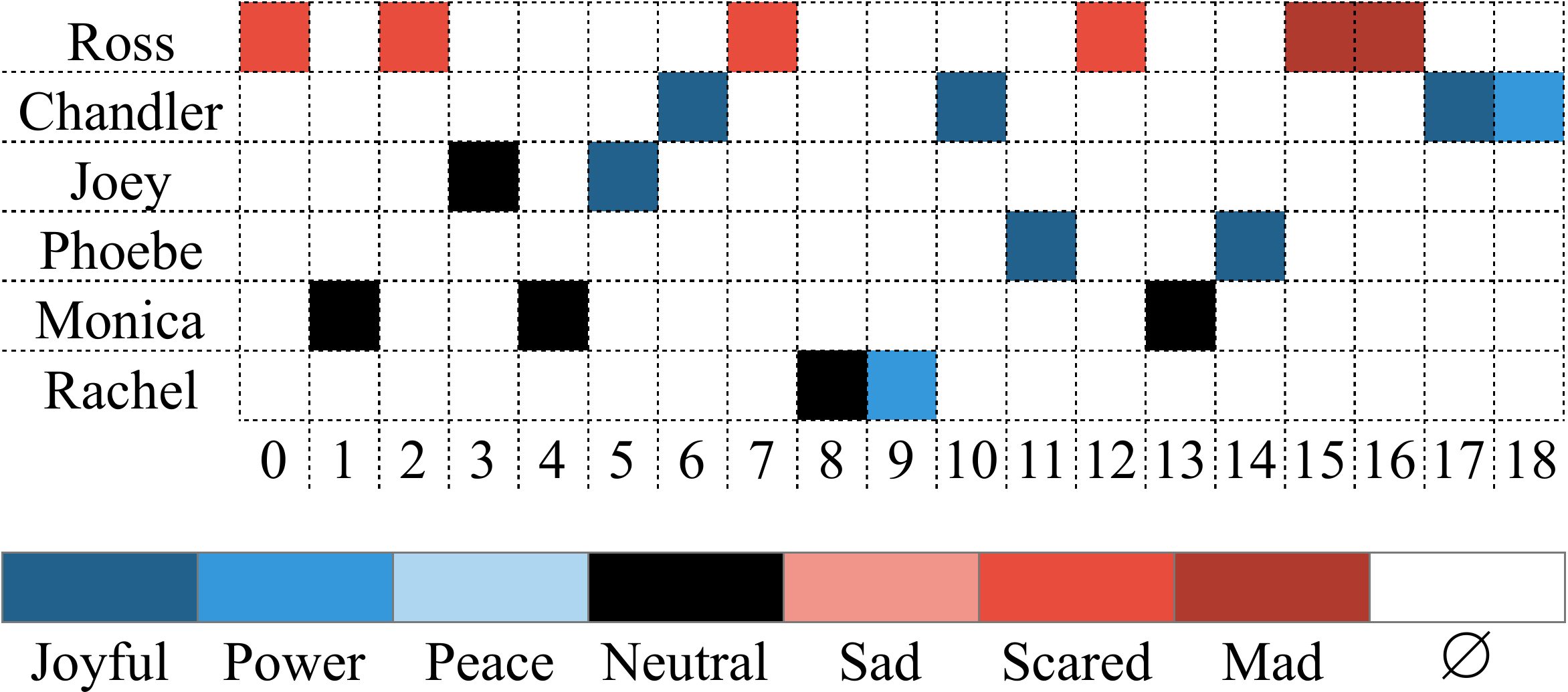}
\caption{Emotions of the main characters within a scene. Rows correspond to the main characters' emotions and columns show the utterance number. No talking is occurred in the white regions.}
\label{fig:dial_seq}
\end{figure}

\noindent Figure~\ref{fig:dial_seq} illustrates how emotions of the six main characters are progressing and getting affected by other utterances as time elapses within a scene. 
It is clear that the emotion of the current speaker is often affected by the same speaker's previous emotions as well as previous emotions of the other speakers participating in the dialogue.

\begin{figure}[htbp!]
\includegraphics[width=\columnwidth]{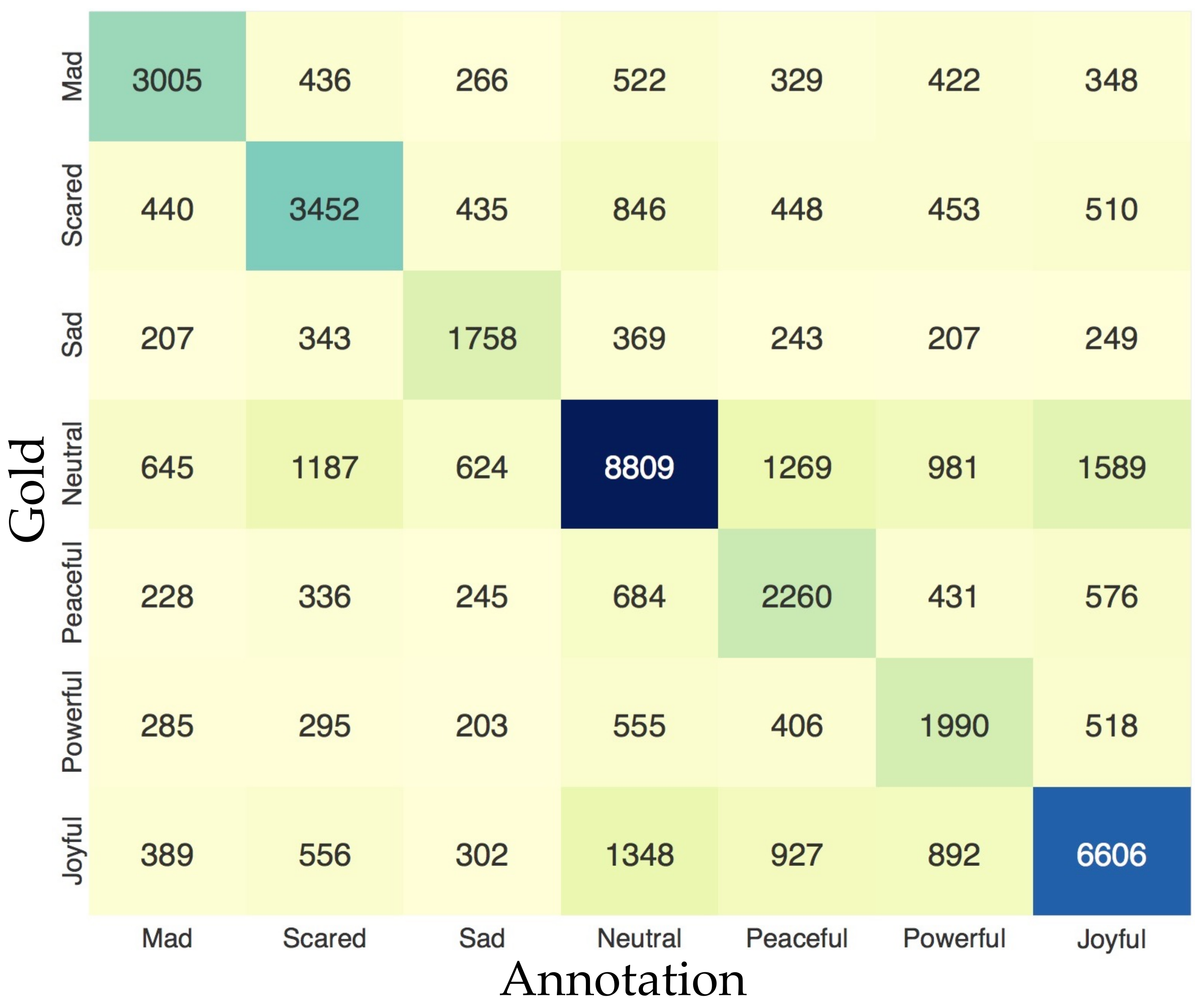}
\caption{Confusion matrix of corpus annotation. Each matrix cell contains the raw count.}
\label{fig:conf_dataset}
\end{figure}

\noindent Figure~\ref{fig:conf_dataset} shows the confusion matrix with respect to the annotation (dis)agreement. Rows correspond to the labels obtained using the voting scheme (Section~\ref{ssec:voting-ranking}), and columns represent the emotions selected by each of four annotators.
The two dominant emotions, \textit{neutral} and \textit{joyful}, cause the most confusion for annotators whereas the minor emotions such as \textit{sad}, \textit{powerful}, or \textit{peaceful} show good agreement on the diagonal.

\subsection{Comparison}

%The size of our corpus is larger than most of the publicly available corpora such as the ISEAR databank,\footnote{\url{emotion-research.net/toolbox/toolboxdatabase.2006-10-13.2581092615}} the SemEval'07 Task 14 ~\cite{strapparava2007semeval}, and the WASSA'17 Task 1~\cite{MohammadB17wassa} datasets.
The ISEAR databank consists of 7,666 statements and six emotions,\footnote{\url{emotion-research.net/toolbox/toolboxdatabase.2006-10-13.2581092615}} which were gathered from a research conducted by psychologists on 3,000 participants.
The SemEval'07 Task 14 dataset was created from news headlines; it contained 250 sentences annotated with six emotions~\cite{strapparava2007semeval}.
The WASSA'17 Task 1 dataset was collected from 7,097 tweets and labeled with four emotions~\cite{MohammadB17wassa}.
The participants of this shared task were expected to develop a model to detect the intensity of emotions in tweets.
Our corpus is larger than most of the other text-based corpora and the only kind providing emotion annotation on dialogue sequences conveyed by consecutive utterances.
%Our corpus contains 7 classes and is made from spoken dialogue.

\begin{figure*}[htbp!]
\centering\small
\includegraphics[scale=0.35]{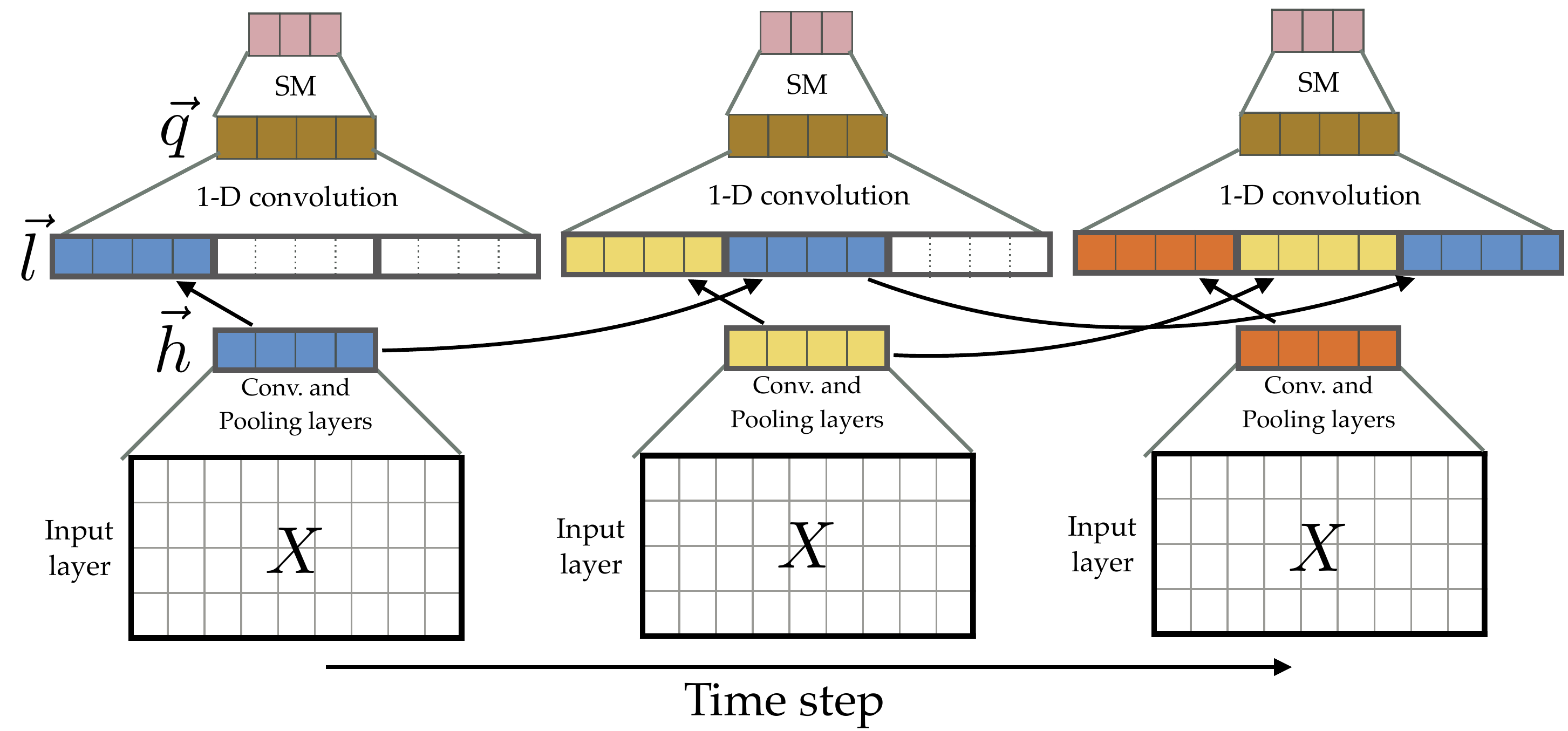}
\caption{The overview of the sequence-based CNN using concatenation ({\textit{SCNN\bm{$_c$}}}), SM: Softmax.}
\label{fig:scnn}
\vspace{-2ex}
\end{figure*}

\section{Sequence-Based Convolutional Neural Networks (SCNN)}
\label{approach}
A unique aspect about our corpus is that it preserves the original sequence of utterances given each dialogue such that it allows to tackle this problem as a sequence classification task.
This section introduces sequence-based CNN models that utilize the emotion sequence from the previous utterances for detecting the emotion of the current utterance.
Additionally, attention mechanisms are suggested for better optimization of these SCNN models.
%The models we will describe in this section are based on the CNN model introduced by \newcite{kim2014convolutional}. 

\subsection{Sequence Unification: Concatenation}
\label{1cnn}

We present a Sequence-based CNN that leverages the sequence information to improve classification.
Figure~\ref{fig:scnn} describes our first SCNN model.
The input to SCNN is an embedding matrix $X\in\mathbb{R}{^{t\times m}}$, where $t$ is the maximum number of tokens in any utterance and $m$ is the embedding size.
Each row in $x$ represents a token in the utterance.
At each time step, region sizes of $\rho\in \{1,2,3,...,r\}, r \in\mathbb{N}$ are considered. %\footnote{To some extent, they work similarly to the \textit{n}-gram filters.} 
Each region has the $f$-number of filters.
As a result of applying convolution and max-pooling, a univariate feature vector $\vec{h}\in\mathbb{R}{^{r\cdot f}}$ is generated for the current utterance.

In the next step, the dense feature vectors from the current utterance and $k$-1 previous utterances within the same dialogue get concatenated column-wise.
As a result of this concatenation, the vector $\vec{l}\texttt{}\in\mathbb{R}{^{r\cdot f\cdot k}}$ is created.
Then, 1-D convolution is applied to $\vec{l}$ to obtain the vector $\vec{q}\in\mathbb{R}{^{(r\cdot f\cdot k-F)/S +1}}$, where $S$ is the stride and $F$ is the receptive field of the 1-D convolution. As a result of this operation, features extracted from the current utterance gets fused with the features associated with the previous utterances. In the final step, softmax is applied for the classification of seven emotions.

%each utterance is decomposed into tokens, where each token is represented by its embedding

\subsection{Sequence Unification: Convolution}
\label{2cnn}

\begin{figure}[htbp!]
\includegraphics[width=\columnwidth]{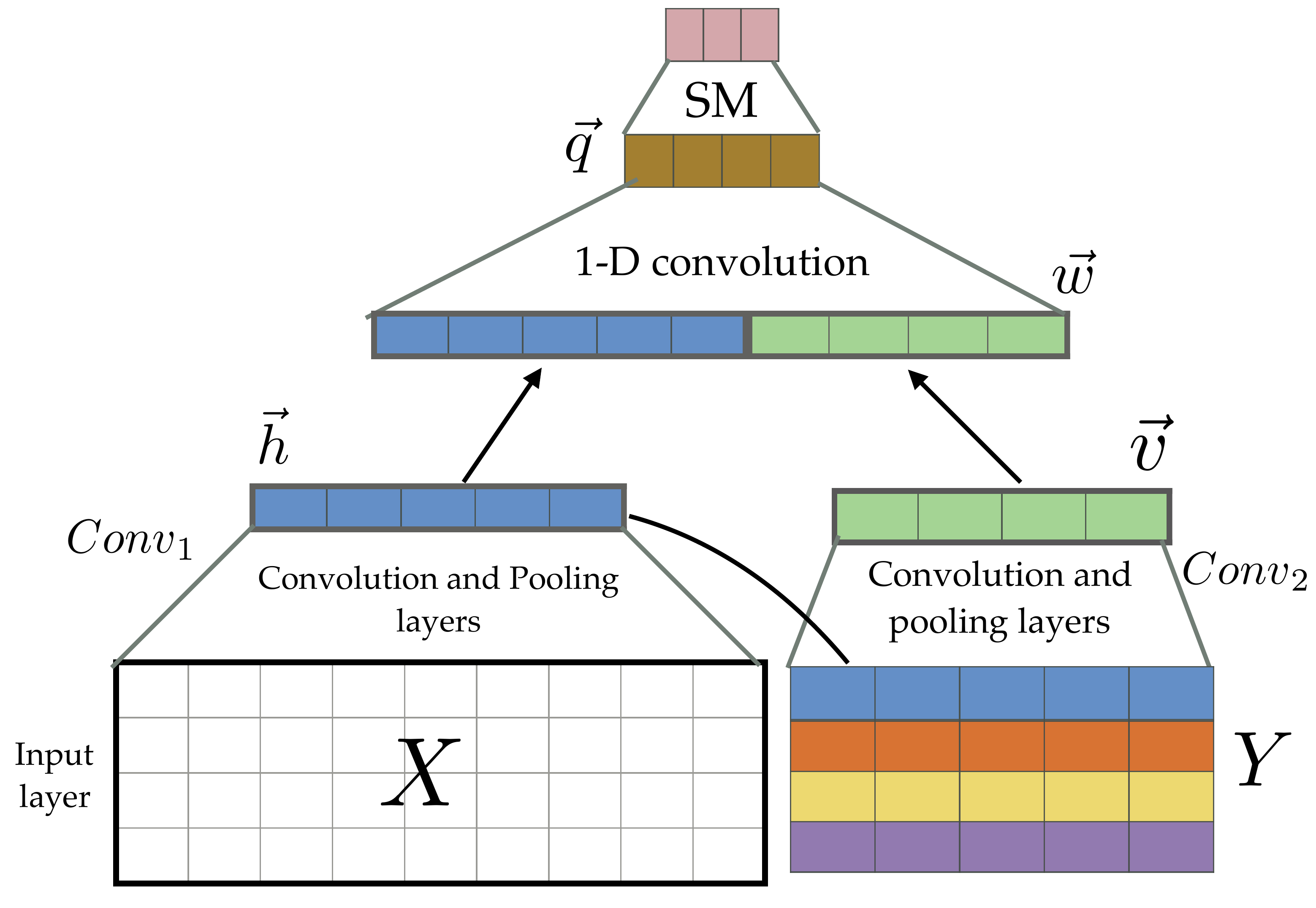}
\caption{The {\textit{SCNN\bm{$_v$}}} model, SM: Softmax.}
\label{fig:scnn3}
\vspace{-1.5ex}
\end{figure}

Let us refer the model in Section~\ref{1cnn} to \textbf{\textit{SCNN\bm{$_c$}}}.
In the second proposed model, referred to \textbf{\textit{SCNN\bm{$_v$}}} (Fig. \ref{fig:scnn3}), two separate 2-D convolutions, Conv$_1$ and Conv$_2$, are utilized for sequence unification.
The input to Conv$_1$ is the same embedding matrix $X$.
The input to Conv$_2$ is another matrix $Y\in\mathbb{R}{^{k\times r\cdot f}}$, which is a row-wise concatenation of the dense vectors generated from the Conv$_1$ of the current utterance as well as $k$-1 previous utterances.
Conv$_2$ has region sizes of $\beta\in\{1,..,b\},b\in\mathbb{N}$ with the $d$-number of filters for each region size.
The output of Conv$_2$ is the vector $\vec{v}\in\mathbb{R}^{d\cdot b}$.
Conv$_1$ and Conv$_2$ are conceptually identical although they have different region sizes, filters and other hyper-parameters.

\noindent In the next step, the outputs of two convolutions get concatenated column-wise to create the vector $\vec{w}\in\mathbb{R}^{(r\cdot f+d\cdot b)}$, which is fed into an one-dimensional CNN to create the vector $\vec{q} \in \mathbb{R}^{(r\cdot f+d\cdot b-F)/S + 1}$, that is the fused version of the feature vectors. 
Finally, the vector $\vec{q}$ is passed to a softmax layer for classification. 
Note that the intuition behind Conv$_2$ is to capture features from the emotion sequence as Conv$_1$ captures features from $n$-grams.

%\noindent Figure \ref{fig:scnn3} shows the overview of this model. In this paper we use  \textbf{\textit{SCNN\bm{$_v$}}} notation to refer to this model.   
%In this paper we use \textbf{\textit{SCNN\bm{$_c$}}} to refer to the SCNN that the feature vector belongs to the current utterance is concatenated with \textbf{\textit{k-1}} feature vectors generated from \textbf{\textit{k-1}} previous utterances (they all belong to the same scene).  

\begin{figure*}[htbp!]
\centering\small
\includegraphics[width=\textwidth]{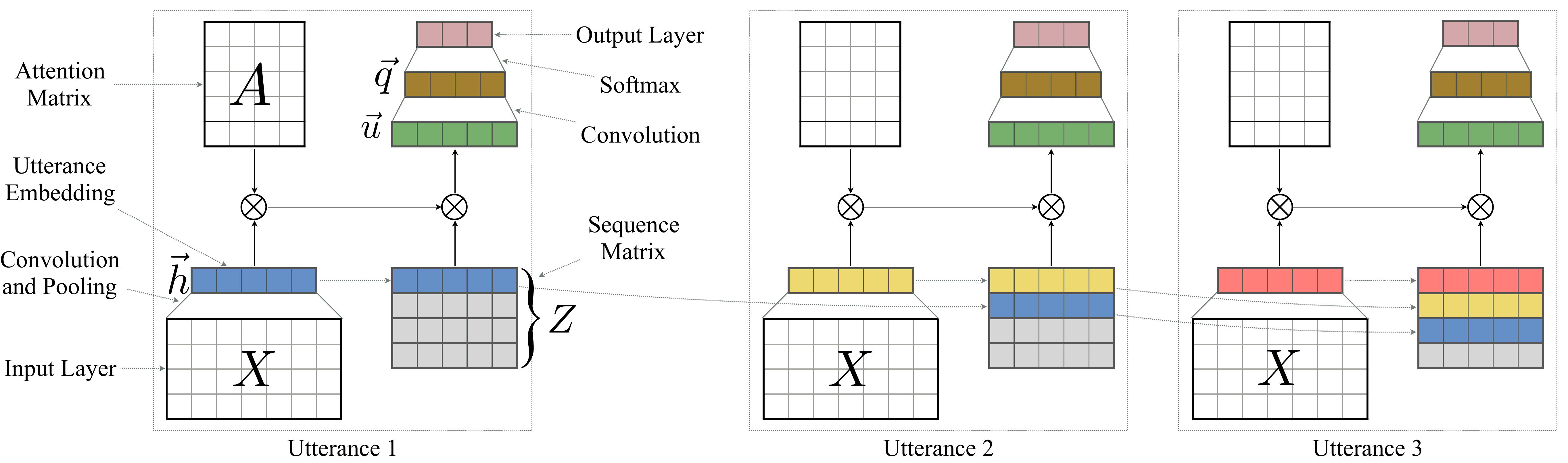}
\caption{The overview of {\textit{SCNN\bm{$_c^a$}}} model.}
\label{fig:scnn2a}
\end{figure*}

\subsection{Attention Mechanism}
\label{ssec:attention}

\noindent We also equipped the {\textit{SCNN\bm{$_c$}}} and {\textit{SCNN\bm{$_v$}}} models with an attention mechanism.
This mechanism allows the SCNN models to learn what part of the features should be attended more.
Essentially, this attention model is a weighted arithmetic sum of current utterance's feature vector where the weights are chosen based on the relevance of each element of that feature vector given the unified feature vectors from the previous utterances.

Figure \ref{fig:scnn2a} depicts attention on {\textit{SCNN\bm{$_c$}}}, \textbf{\textit{SCNN\bm{$_c^a$}}}.
In this model, the current feature vector $\vec{h}$ and the $k$-1 previous feature vectors get concatenated row-wise to create $Z\in\mathbb{R}^{k\times r\cdot f}$.
An attention matrix $A\in\mathbb{R}^{r\cdot f\times k}$ is applied to the current feature vector.
The weights of this attention matrix are learned given the past feature vectors.
The result of this operation is $\vec{u}=\vec{h} \times A \times Z, \vec{u}\in\mathbb{R}^{1 \times rf}$.
Finally, 1-D convolution and softmax are applied to $\vec{u}$.

\begin{figure}[htbp!]
\includegraphics[width=\columnwidth,height=7cm]{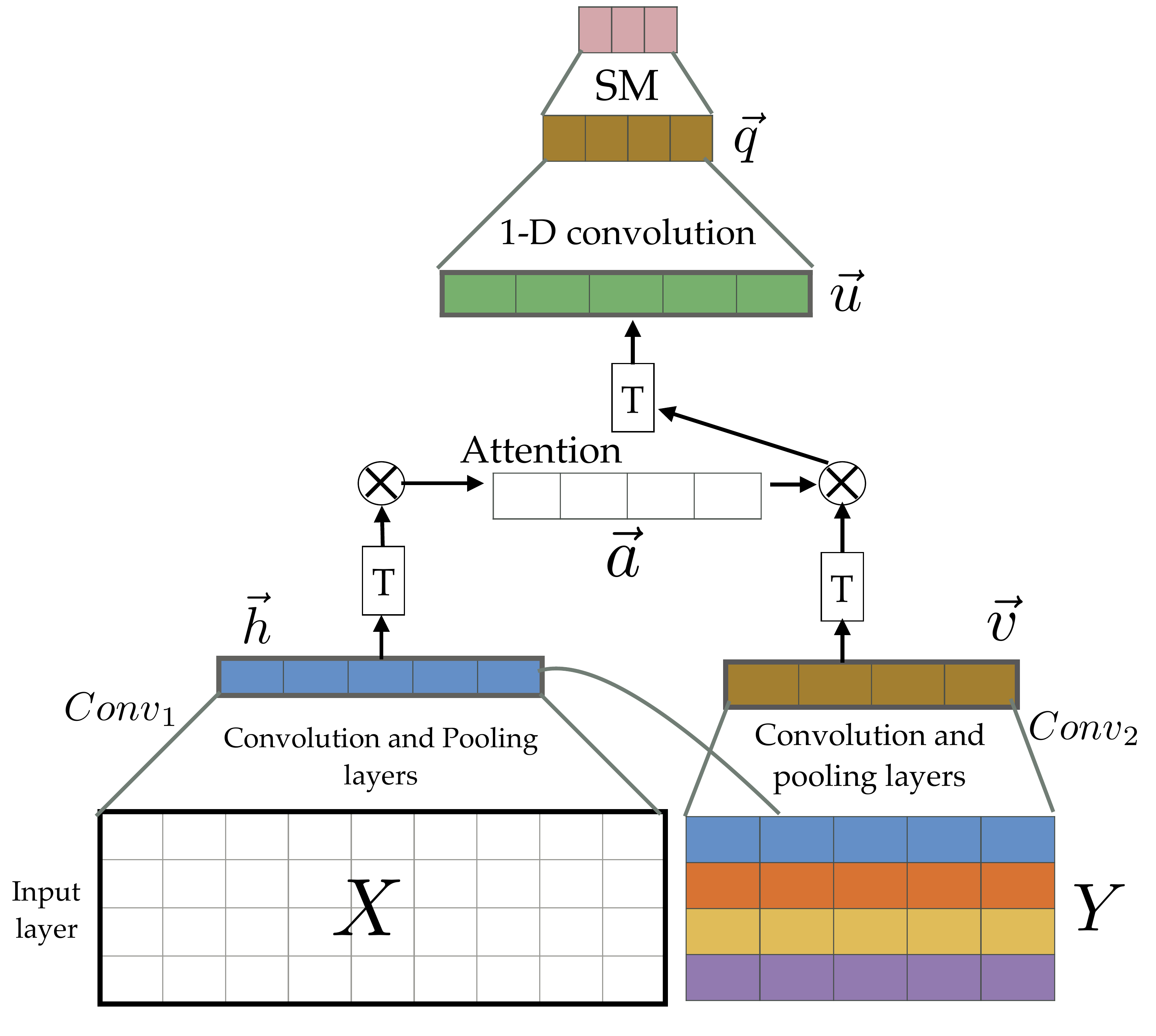}
\caption{The {\textit{SCNN\bm{$_v^a$}}} model, SM:Softmax.}
\label{fig:scnn3a}
\end{figure}

\noindent Figure \ref{fig:scnn3a} shows another attention model, called \textbf{\textit{SCNN\bm{$_v^a$}}}, based on \textbf{\textit{SCNN\bm{$_v$}}}. 
In this model, an attention vector $\vec{a}$ with trainable weights is applied to the outputs of Conv$_1$ ($\vec{h}$) and Conv$_2$ ($\vec{v}$).
The output of the attention vector is a vector $\vec{u} = (h ^{T} \times \vec{a} \times v ^{T})^{T}, \vec{a} \in \mathbb{R}^{1 \times d \cdot b}$. 
Finally, 1-D convolution and softmax are applied to $\vec{a}$ to complete the classification task. 
The multiplication sign in Figures \ref{fig:scnn2a} and \ref{fig:scnn3a} are the matrix multiplication operator and the \bm{$T$} sign refers to the transpose operation.

Generally the inputs to attention mechanism have a fixed size whereas in our model one of the input comes from dynamically generated embedding of previous hidden layers and the other input is the dense representation of the current utterance.

\section{Experiments}
\label{exp}

%In this section we quantitatively compare our proposed models with the based models. We will show that the best result would be achieved with \textbf{\textit{SCNN\bm{$_c^a$}}} model.

\subsection{Corpus}
Our corpus is split into training, development, and evaluation sets that include 77, 11, and 9 episodes, respectively.
Although episodes are randomly assigned, all utterances from the same episode are kept in the same set to preserve the sequence information.
Further attempts are made to maintain similar ratios for each emotion across different sets.
Table \ref{sets_dist} shows the distributions of the datasets.

\begin{table}[htbp!]
\centering
\resizebox{\columnwidth}{!}{
\begin{tabular}{c||r|r|r|r|r|r|r||r}
\bf Set & \multicolumn{1}{c|}{\bf N} & \multicolumn{1}{c|}{\bf J} & \multicolumn{1}{c|}{\bf P} & \multicolumn{1}{c|}{\bf W} & \multicolumn{1}{c|}{\bf S} & \multicolumn{1}{c|}{\bf M} & \multicolumn{1}{c||}{\bf D} & \multicolumn{1}{c}{\bf Total} \\
\hline\hline
TRN & 3,034 & 2,184 & 899 & 784 & 1,286 & 1,076 & 671 & 9,934 \\%(78.80) \\
DEV &   393 &   289 & 132 & 134 &   178 &   143 &  75 & 1,344 \\%(10.66) \\
TST &   349 &   282 & 159 & 145 &   182 &   113 &  98 & 1,328 \\%(10.54) \\
\end{tabular}}
\caption{The number of utterances in each dataset. N: neutral, J: joyful, P: peaceful, W: powerful, S: scared, M: mad, D: sad.}
\label{sets_dist}
\end{table}

\begin{table*}[htbp!]
\centering\small
\begin{tabular}{l||r|r||r|r||r|r||r|r}
 & \multicolumn{2}{c||}{\bf SCNN \bm{$_c$}} & \multicolumn{2}{c||}{\bf SCNN \bm{$_v$}}  & \multicolumn{2}{c||}{\bf SCNN \bm{$_c^a$}} & \multicolumn{2}{c}{\bf SCNN \bm{$_v^a$}}\\ \cline{2-9}
       & \bf Acc\bm{$_7$} &  \bf F1\bm{$_7$} & \bf Acc\bm{$_7$} & \bf F1\bm{$_7$} & \bf Acc\bm{$_7$} & \bf F1\bm{$_7$} & \bf Acc\bm{$_7$} & \bf F1\bm{$_7$} \\  \hline\hline
1    &     38.30    &     24.1    &     36.5   &     21.00   &     38.30   &     26.60    &     38.10   &     25.70 \\
2   &     38.80    &     23.0    &     37.01   &     20.00   &     38.60    &     28.00    &     38.70   &     25.60 \\ 
3 &     \bf 39.10    &   \bf  25.74    &     37.00   &     21.00   &     \bf 39.70    &   \bf  28.50    &     38.46   &  26.15 \\
4 &     37.40    &     22.35    &     36.30   & 20.00   &     38.00    &     26.80    &     38.16   &     25.20 \\
5 &    38.20  &  24.1   &  \bf 37.20   &    \bf 22.00   &  38.90   & 27.76    &  \bf 38.90 &     \bf 28.20 

\end{tabular}
\caption{The performance of different models (in \%) on development set. Acc\bm{$_7$}: accuracy for 7 classes, F1\bm{$_7$}: macro-average F1-score for 7 classes. First column refers to number of sequence information.}
\label{results_dev}
\end{table*}

\subsection{Preprocessing }

In this work we utilize Word2vec word embedding model introduced by \newcite{mikolov2013distributed}. The word embedding is trained separately using Friends TV show transcripts, Amazon reviews, New York times and Wall street journal articles. In this research we use word embedding size of 200.% as we achieved the best performance on the development set with this embedding size.   

%Each utterance is consist of sentence(s).  Words existing in each utterance should be tokenized first and later vectorized. The vectorized representation of tokens will be then fed into the models we will explain in this section.

\subsection{Models}

We report the results of the following four models: \textbf{\textit{SCNN\bm{$_c$}}} , \textbf{\textit{SCNN\bm{$_c^a$}}}, \textbf{\textit{SCNN\bm{$_v$}}} and \textbf{\textit{SCNN\bm{$_v^a$}}}. For each of the mentioned models, we collect the results using $n\in\{1,...,5\}$ previous utterances. For a better comparison we also report the results achieved from base CNN \cite{kim2014convolutional} and RNN-CNN. RNN-CNN is our replication of the model proposed by \newcite{donahue2015long} to fuse time-series information for visual recognition and description.
RNN-CNN that we employ is comprised of a Long Short Term Memory (LSTM); the input to LSTM is the feature vectors generated by CNN for all the utterances in a scene and we train RNN-CNN to tune all the hyper-parameters.

\subsection{Results}

%SCNN$_1$

\noindent Table \ref{results_dev} summarizes the overall performance of our proposed sequence-based CNN models on the development set. First column of the table indicates the number of previous utterances included in our model. We report both accuracy and F1-score for evaluating the performance of our models. F1-score, by considering false positives and false negatives is generally a better way of measuring the performance on a corpus with unbalanced distributed classes. In \textbf{\textit{SCNN\bm{$_c$}}} and \textbf{\textit{SCNN\bm{$_c^a$}}}, by including previous three utterances, and in  \textbf{\textit{SCNN\bm{$_v$}}} and \textbf{\textit{SCNN\bm{$_v^a$}}}, by considering  previous five utterances the best results are achieved. We also ran our experiments with considering more than five utterances, however no significant improvement was observed.

%for example third row of the table shows the performance of four proposed models using only three previous utterances' information

\begin{table}[htbp!]
\centering\small

\begin{tabular}{l||r|r|r|r}
 & \multicolumn{4}{c}{\bf Evaluation}  \\ \cline{2-5}
       & \bf Acc\bm{$_7$} & \bf Acc\bm{$_3$} & \bf F1\bm{$_7$} & \bf F1\bm{$_3$}  \\  \hline\hline
CNN    &     37.01    &     49.78    &     22.91  &     36.83   \\
RNN-CNN   &     29.00    &     42.10    &     11.00   &     24.05    \\ \hline
SCNN {$_c$} &     37.35    &     53.20   &     25.06   &     38.00   \\
SCNN {$_v$} &     36.45    &     51.11    &     21.00   & 36.50    \\
SCNN {$_c^a$} & \bf 37.90    & \bf 54.00    & \bf 26.90   &    \bf 39.25   \\
SCNN {$_v^a$} &     37.67    &     51.90    &     26.70   &     38.21  
\end{tabular}\

\caption{The performance of different models (in \%) on evaluation set. Acc\bm{$_7$}: accuracy for 7 classes, Acc\bm{$_3$}: accuracy for 3 classes (Table~\ref{tbl:emotion-dist}), F1\bm{$_7$}: macro-average F1-score for 7 classes, F1\bm{$_3$}: macro-average F1-score for 3 classes.}
\label{results_eval}
\end{table}

Table \ref{results_eval} summarizes the overall performance of our models on the evaluation set. To compare our proposed models to some other baseline models, we also include performances of CNN and RNN-CNN. The accuracies and the F1-scores are reported for the cases of 7 and 3 emotions (Table~\ref{tbl:emotion-dist}), where the latter case is comparable to the typical sentiment analysis task. For the models listed in table \ref{results_eval}, we choose the sequence numbers that had the best performances on the development set (3 for {\textit{SCNN\bm{$_c$}}} and {\textit{SCNN\bm{$_c^a$}}}, 5 for {\textit{SCNN\bm{$_v$}}} and {\textit{SCNN\bm{$_v^a$}}}). From tables \ref{results_dev} and \ref{results_eval}, we can see that \textbf{\textit{SCNN\bm{$_c^a$}}} outperformed all other listed models.

To fuse the generated dense feature vectors we applied different combinations of regularized feature fusion networks similar to the networks employed in \newcite{wu2015modeling} and \newcite{bodla2017deep}. However, for our task, none of these fusion networks performed better than the 1-D convolutional layer we utilized.
It worth to be mention that, in the first time step of our both attentive models, the two inputs to the attention matrix/vector are two identical vectors (first utterance's feature vector) which mean that the main impact of attention mechanism starts from the second time step.

\subsection{Analysis}

\begin{figure}[htbp!]

\includegraphics[width=\columnwidth,height=7cm]{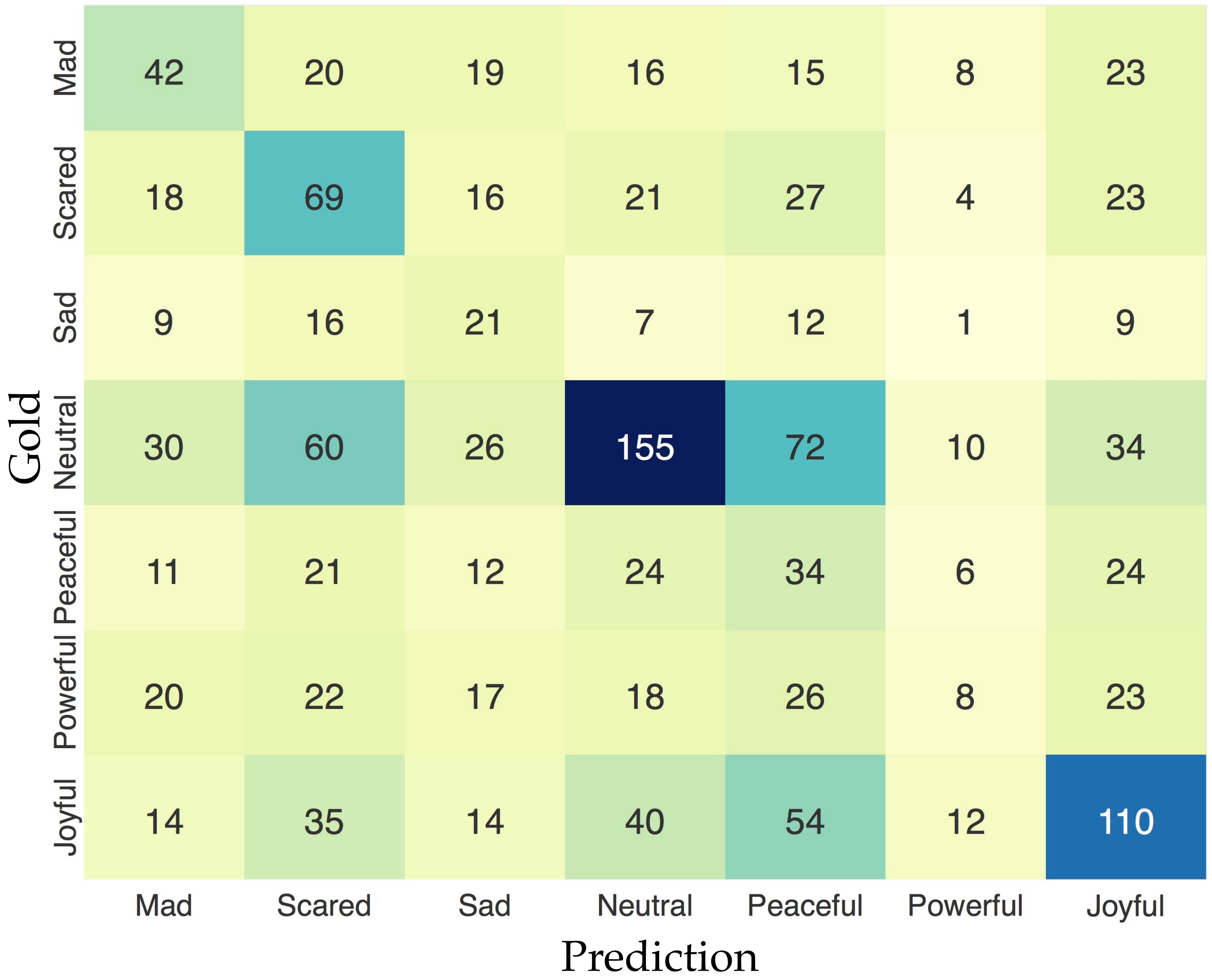}
\caption{Confusion matrix of the best model on the evaluation set. Each matrix cell contains the raw count.}
\label{fig:conf_pred}
\end{figure}

Figure \ref{fig:conf_pred} shows the confusion matrix of gold labels and prediction for our best model,  {\textit{SCNN\bm{$_c^a$}}}. Mostly all of the emotions get confused the most with \textit{neutral}. \textit{Peaceful} has the highest rate of confusion with \textit{neutral}; 30\% of total number of examples in this class are confused with \textit{neutral}. Whereas, \textit{joyful} and \textit{powerful} have the least confusion rates with \textit{neutral} (13.8\% and 20.4\% respectively).

\noindent To further explore the effect of sequence number, we divided the evaluation set into four batches based on the number of utterances in the scenes as described in section \ref{crowd}. After examining the performances of  {\textit{SCNN\bm{$_c$}}} and {\textit{SCNN\bm{$_c^a$}}} on four batches, we noticed these two models by considering previous three sequences, performed better (roughly 4\% boost in F1-scores) on the first two batches (i.e. scenes containing [5,15) utterances) compared to other models such as base CNN. It seems, in very long scenes which usually contain more speakers and transitions between the speakers, our proposed models did not significantly outperform base CNN.

During our experiments we observed that RNN-CNN did not have a compelling performance. Generally, complicated models with higher number of hyper parameters require a larger corpus for tuning the hyper parameters. Given the relatively small size of our proposed corpus for such a model, we noticed that RNN-CNN over fitted rapidly at the very early epochs. It was basically well-tuned to detect two most dominate classes. Also, \textbf{\textit{SCNN\bm{$_v$}}}, did not outperform base CNN model although it did outperform base CNN after including attention mechanism ({\textit{SCNN\bm{$_v^a$}}}). We believe that size of our corpus could be inadequate to train this model which has more hyper-parameters than {\textit{SCNN\bm{$_c$}}}.

\begin{figure}[htbp!]
\includegraphics[width=\columnwidth]{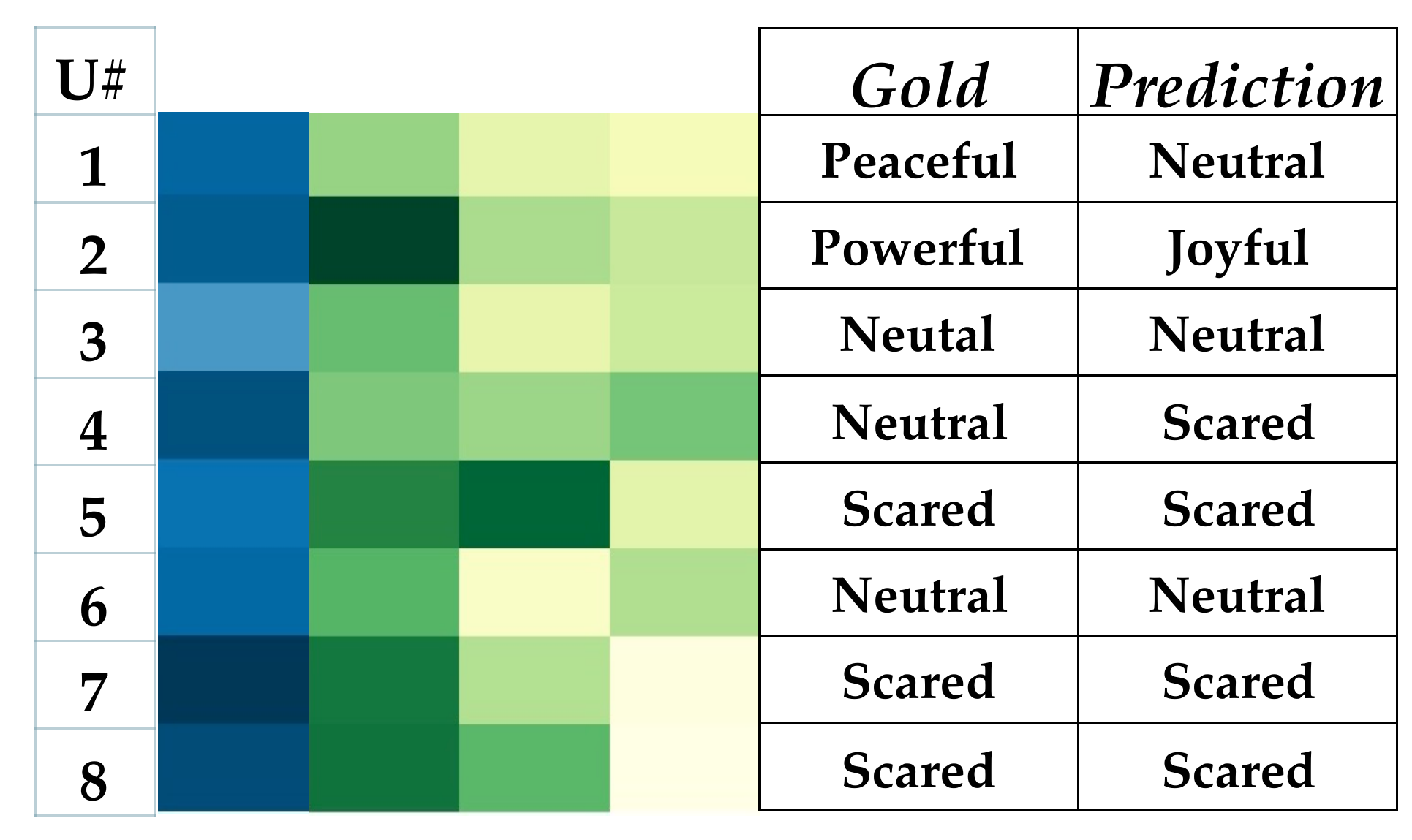}
\caption{ The heat-map representation of $\vec{h}\times A$ in \textit{SCNN\bm{$_c^a$}} model, U\#: Utterance number.}
\label{fig:atten}
\end{figure}

Figure \ref{fig:atten} depicts the heat-map representation of the vector created from multiplication of current utterance's feature vector with the attention matrix $\textbf{A}$, in \textit{SCNN\bm{$_c^a$}}. The heat-map includes the first eight consecutive utterances (rows of the heat-map) of a scene. 
Each row shows importance of the current utterance (color-mapped with blue) as well as previous three utterances (last three columns of the heat-map) at that particular time step. At each time step, the current utterance has the largest value which means our model attends more to the current utterance compared to previous ones. Attention matrix learns to assign the weights to previous utterances given the weight assigned to the current utterance.  

\noindent For instance, utterance number 7 and 8 attend more to their previous two utterances (which have similar emotions). Utterance number 5 attends less to utterance 2 as the second utterance has a positive emotion. Similarly, utterance 2 attends more to its previous utterance as they both have a positive emotion. Generally, in the cases where the emotion of current utterance is \textit{neutral}, the weights assigned to its previous utterances are relatively small and mostly similar.

%At time step \textit{k}, the first column of the heat-map is created from $\vec{h}{_k} \times A$ where $\vec{h}{_k} $ is current utterance's feature vector, similarly second column belongs time step \textit{k}, second column of the heat-map refers to the utterance at time step \textit{k-1} and so on. The first columns of the heat-maps have the larges values which refer to the current utterances. 

\section{Conclusion}

In this work, we introduced a new corpus for emotion detection task which was gathered from the spoken dialogues. We also proposed attentive SCNN models which incorporated the existing sequence information. The experimental results showed that   our proposed models outperformed the base CNN. As annotating emotions from text are usually subjective, in future, we plan to assign more annotators to improve the quality of the current corpus. Also, to fully evaluate the performances of our proposed models we intend to implement different combinations of attention mechanisms and expand the size of our corpus by annotating more seasons of Friends TV show.

\bibliography{ijcnlp2017}
\bibliographystyle{ijcnlp2017}

\end{document}